\documentclass{article}

\usepackage[preprint,nonatbib]{neurips_2023}

\usepackage[utf8]{inputenc} 
\usepackage[T1]{fontenc}    
\usepackage{url}            
\usepackage{booktabs}       
\usepackage{amsfonts}       
\usepackage{nicefrac}       
\usepackage{microtype}      
\usepackage[usenames,dvipsnames]{xcolor} 
\usepackage{graphicx}
\usepackage{multirow}
\usepackage{float}

\usepackage[backend=biber, style=ieee, citestyle=numeric-comp, dashed=false]{biblatex}

\definecolor{citecolor}{HTML}{0071BC}
\definecolor{linkcolor}{HTML}{ED1C24}
\definecolor{highlightcolor}{HTML}{ABCDEF}
\usepackage[breaklinks, colorlinks, citecolor=citecolor, linkcolor=linkcolor, bookmarks=false]{hyperref}

\title{CVPR 2023 Text Guided Video Editing Competition}

\author{%
  {\bf Jay Zhangjie Wu$^{2*}$, Xiuyu Li$^{1*}$, Difei Gao$^{2*}$, Zhen Dong$^{1*}$, Jinbin Bai$^{2*}$, Aishani Singh$^{1*}$,} \\
  {\bf Xiaoyu Xiang$^{3*}$, Youzeng Li$^{4\dag}$, Zuwei Huang$^{4\dag}$, Yuanxi Sun$^{4\dag}$, Rui He$^{4\dag}$, Feng Hu$^{4\dag}$,} \\
  {\bf Junhua Hu$^{4\dag}$, Hai Huang$^{4\dag}$, Hanyu Zhu$^{4\dag}$, Xu Cheng$^{4\dag}$, Jie Tang$^{4\dag}$,} \\
  {\bf Mike Z. Shou$^{2*}$, Kurt Keutzer$^{1*}$, Forrest N. Iandola$^{3*}$}
  \\
  $^1$ University of California, Berkeley \\
  $^2$ National University of Singapore \\
  $^3$ Meta \\
  $^4$ Tencent Holdings Ltd and Tsinghua University \\
  $^*$ Competition organizer \\
  $^\dag$ Competition winner
}

\begin{document}

\maketitle

\begin{abstract}
Humans watch more than a billion hours of video per day\footnote{https://www.comparitech.com/tv-streaming/youtube-statistics/}.
Most of this video was edited manually, which is a tedious process.
However, AI-enabled video-generation and video-editing is on the rise.
Building on text-to-image models like Stable Diffusion and Imagen, generative AI has improved dramatically on video tasks.
But it's hard to evaluate progress in these video tasks because there is no standard benchmark.
So, we propose a new dataset for text-guided video editing (TGVE), and we run a competition at CVPR to evaluate models on our TGVE dataset.
In this paper we present a retrospective on the competition and describe the winning method.
The competition dataset is available at \url{https://sites.google.com/view/loveucvpr23/track4}.

\end{abstract}

\section{Introduction}
\label{sec_intro}
Leveraging AI for video editing has the potential to unleash creativity for artists of all skill levels.
In the last year, numerous papers have been written on Text Guided Video Editing (TGVE), including Dreamix~\cite{dreamix}, Tune-A-Video~\cite{tune_a_video},  Gen-1~\cite{gen1}, among others~\cite{shin2023edit, videoP2P, wang2023zero, qi2023fatezero, ceylan2023pix2video, khachatryan2023text2videozero, yang2023rerender, geyer2023tokenflow, zhao2023motiondirector, liu2023dynvideo}.
While the qualitative results are getting more and more impressive, there is no standard quantitative benchmark in this field.
Without a standard benchmark, and with many papers being closed-source, it is impossible to know which model is the state-of-the-art or to understand the strengths and weaknesses of different models. 

To address this, we have made these contributions:
\begin{itemize}
    \item Released an open-source dataset of 76 videos, each with 4 prompts for text guided video editing
    \item Organized a competition workshop at CVPR 2023 to evaluate TGVE models 
    \item Conducted human evaluation and automated evaluations to rank 8 TGVE models
\end{itemize}

In this paper, we will describe the dataset, the evaluation methodology, and the findings from the competition.
We will also describe the model that won the competition.

\begin{figure}
  \centering
  \includegraphics[width=1\linewidth]{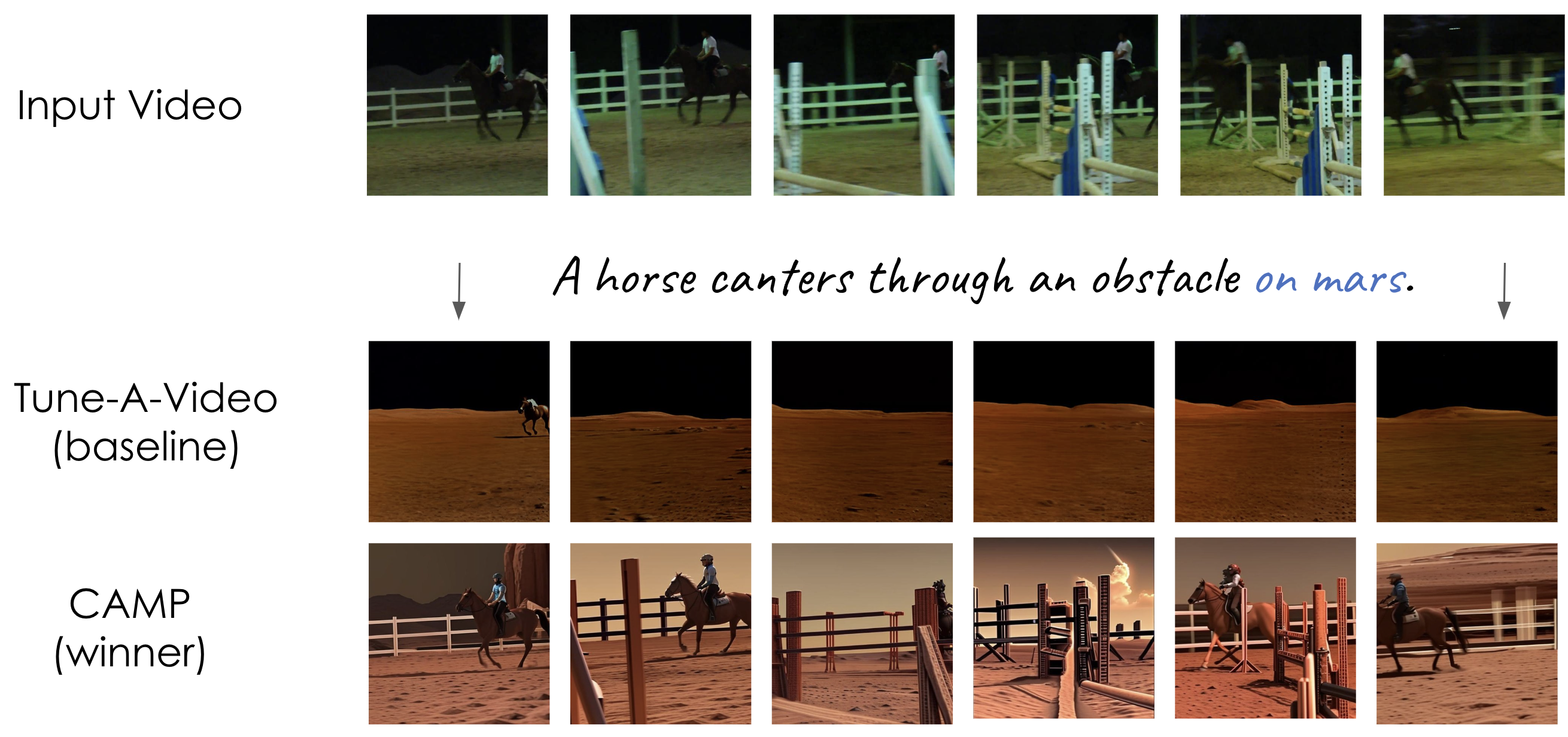}
  \caption{Example background-change task in the TGVE 2023 competition.}
   \label{fig:horse-jump-canter}
\end{figure}

\section{Related Work}

The development of text-to-image models such as Stable Diffusion has led to rapid process in computer-generated art, including text-to-3D~\cite{dreamfusion}, text-to-audio~\cite{audioldm}, to text-to-video~\cite{imagen_video, zhang2023show1, makeavideo, align}.
While text-to-video is an incredibly challenging problem for today's generative AI technology, it is our view that generative AI will become adept at {\em editing existing videos} over the next couple of years.
Five years ago, using AI to edit a video in any complex way (e.g. anything more complicated than a simple style-transfer) was nearly unthinkable. 
However, building on top of advances in text-to-image models, works such as Dreamix~\cite{dreamix}, Tune-A-Video~\cite{tune_a_video}, and Gen-1~\cite{gen1} are able to take a video and a text prompt and edit the video to match the prompt.

In terms of algorithms, there are many options for how to design a TGVE model. 
Tune-A-Video~\cite{tune_a_video} is the first open-source video diffusion model for TGVE. It inflates an image diffusion model into a video model with cross-frame attention, and finetunes it on a single video to generate videos with related motion. Based on it, Edit-A-Video~\cite{shin2023edit}, Video-P2P~\cite{videoP2P}, vid2vid-zero~\cite{wang2023zero} exploit Null-Text Inversion for precise inversion to preserve the unedited region. Dreamix~\cite{dreamix} instead finetunes a video foundation model~\cite{ho2022imagen} pretrained on large-scale video data, and establishes superior visual quality in TGVE.

Some recent models use zero-shot techniques that don't need to be finetuned on each input video. To preserve structural and motion information in source video, FateZero~\cite{qi2023fatezero} merges attention features pre and post-editing with the editing masks produced by Prompt2Prompt~\cite{hertz2022prompttoprompt}. Text2Video-Zero~\cite{khachatryan2023text2videozero} converts the latent to directly emulate motions, while Pix2Video~\cite{ceylan2023pix2video} aligns the latent of the current frame with previous frames via cross-frame attention. To further enhance pixel-level temporal consistency, Rerender A Video~\cite{yang2023rerender} propagates the edits using temporal-aware patch matching and frame blending. TokenFlow~\cite{geyer2023tokenflow} extracts inter-frame feature correspondences using a nearest-neighbor search and propagates the edited tokens throughout the entire video flow. 

Now, what is a good way to evaluate these models?
In Table~\ref{T_related_datasets}, we summarize the evaluation datasets used in several TGVE papers.
While there isn't a standardized dataset in these papers, there are common themes in the datasets.
The typical setup is 10 to 100 videos, with 2 to 4 prompts per video.
Videos are short (under 10 seconds) and low resolution (often 480x480). 
Common video sources include YouTube and DAVIS~\cite{DAVIS2016, DAVIS2017}.

\begin{table*}[t!]
    \caption{Datasets used for human evaluation in TGVE papers.}
\vspace{1mm}
    \centering
    \begin{tabular}{llcc}
      \toprule
      Model & \# of Eval Videos & Source of Eval Videos & \# of Edit Prompts\\
      \midrule
      Dreamix~\cite{dreamix} & 29 & YouTube & 127 \\
      Gen-1~\cite{gen1} & unknown & DAVIS & 35 \\
      Tune-A-Video~\cite{tune_a_video} & 42 & DAVIS & 140 \\
      Text2LIVE~\cite{text2LIVE} & 7 & DAVIS & unknown \\
      Video-P2P~\cite{videoP2P} & 10 & YouTube & unknown \\
      \textbf{TGVE 2023 (Ours)} & 76 & DAVIS, YouTube, Videvo & 304 \\
      \bottomrule
    \end{tabular}
    \label{T_related_datasets}
\end{table*}

\section{The TGVE 2023 Dataset}
Now we describe our approach for creating the competition dataset. 
We begun by collecting 100 videos from DAVIS, YouTube, and Videvo. 
All of the videos have Creative Commons or other open licenses that allow us to modify and redistribute the videos.
We filtered it down to 76 videos total.
We made the dataset small so that human-evaluation is affordable.

\begin{figure}
  \centering
  \includegraphics[width=1\linewidth]{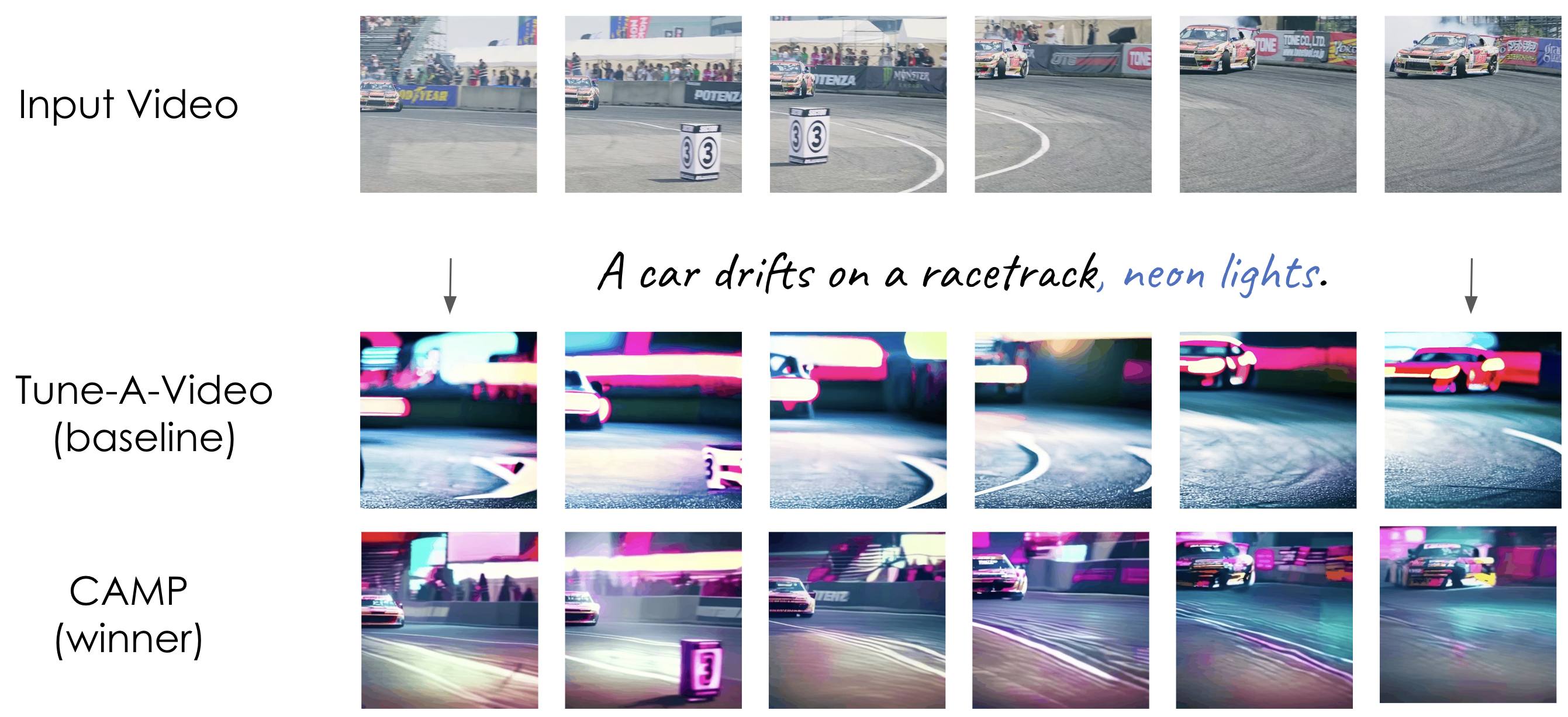}
  \caption{Example style-change task in the TGVE 2023 competition.}
   \label{fig:drift-turn}
\end{figure}

Most of today's AI video-editing methods are only able to handle short, low-resolution videos.
So, instead of using the whole video, we reduce each video to 32 frames (for DAVIS and Videvo) or 128 frames (for YouTube). 
We crop and downscale the videos to 480x480.
In the future, as these methods improve, we may develop a version of the dataset with larger and longer videos.

Each video has a ground-truth caption.
To create the ground-truth captions, we started with the provided caption (e.g. from YouTube), and we manually improved the caption to describe precisely what is happening in the short video clip.

Each video has 4 ``edit captions" that describe how we want the TGVE models to edit the videos.
Specifically, each video has 4 types of edit captions: style-change (e.g. neon lights style), background-change (e.g on the mars), object-change (e.g. change human to panda), or multiple-changes (at least 2 types of edits).
To create the edit captions, we initially asked ChatGPT to take the 76 ground-truth captions and produce edits to the style, background, or objects.
However, the ChatGPT captions were a bit boring -- turning dogs into cats, sunrises into sunsets, and lakes into oceans. 
We manually edited many of the captions to make them more creative -- turning dogs into kangaroos, sunrises into abstract vector-art, and lakes into underwater coral reefs.
We show examples in Figures~\ref{fig:horse-jump-canter} and~\ref{fig:drift-turn}

\subsection{Evaluation Approach}
Most papers on text-guided video editing include both automated metrics (e.g. CLIP score or FVD score~\cite{FVD}) and human metrics (based on people labeling data).
In our conversations with several researchers in this field, everyone said that automated metrics are quite noisy, and human evaluation is more trustworthy.
This matches our own experience.

We used human evaluation to select the winner of the competition. 
Specifically, we developed a mechanical turk interface where labelers are shown 3 videos: the input video, the baseline edited video (the baseline is Tune-A-Video~\cite{tune_a_video}), and the video edited by the proposed model.
After the labeler has watched 3 videos, they need to answer 3 questions:
\begin{itemize}
  \item Text alignment: Which video better matches the caption?
  \item Structure: Which video better preserves the structure of the input video?
  \item Quality: Aesthetically, which video is better?
\end{itemize}
For each submission to the challenge, we used this approach to compare the submitted videos to the Tune-A-Video baseline.

\section{Competition Results}
\label{sec:competition-results}


We received 5 submissions to the competition, and we now summarize the method used in each submission.

{\bf Team PAIR} used the image-based diffusion models ControlNet and InstructPix2Pix~\cite{InstructPix2Pix}. Optical flow is used to help with temporal stability.

{\bf Team RewardT2VE} used Tune-A-Video and Make-A-Protagonist~\cite{makeaprotagonist}. With these methods, the RewardT2VE team generated many videos and used ImageReward~\cite{imagereward} to measure aesthetic quality and CLIPScore~\cite{clipscore} to measure alignment with the prompt. The team submitted the best videos based on these metrics. 

{\bf Team Noah Wukong} used a Video Diffusion Model~\cite{videodiffusionmodel} with 3D convolutions.

{\bf Team T2I\_HERO} used Video-P2P~\cite{videoP2P}. 

{\bf Team CAMP} won the competition with their method, Text-Based Two-Stage Video Editing, which is described later in this paper. 

In Table~\ref{tab:results}, we show the results from all submissions. 
We include automated metrics (CLIPScore~\cite{clipscore} and PickScore~\cite{pickscore}) and human evaluation from Mechanical Turk.
For human evaluation, when we report the number 0.689, it means the human evaluators preferred the proposed method over the baseline method 68.9\% of the time.

We use human evaluation to choose the winners of the challenge, with Team CAMP beating the Tune-A-Video 59.1\% of the time. 
It's interesting to note that the automatic metrics are uncorrelated (perhaps even anti-correlated) with the human evaluation results. 
The contest organizers also reviewed the videos and found that we agreed with human evaluators, and to our eye the CAMP videos were significantly better than the others.

\begin{table}[H]
\caption{Competition results. Human evaluation was used to select the winner.}
\label{tab:results}
\resizebox{\textwidth}{!}{%
\begin{tabular}{lccclcccc}
\toprule
\multirow{2}{*}{Method} &
  \multicolumn{3}{c}{Automatic Metrics (higher is better)} &
   &
  \multicolumn{4}{c}{Human Evaluation (higher is better)} \\ \cline{2-4} \cline{6-9} 
 &
  \begin{tabular}[c]{@{}c@{}}CLIPScore\\ (Text Alignment)\end{tabular} &
  \begin{tabular}[c]{@{}c@{}}CLIPScore\\ (Frame Consistency)\end{tabular} &
  \begin{tabular}[c]{@{}c@{}}PickScore\\ (Aesthetics)\end{tabular} &
   &
  Text Alignment &
  Structure &
  Quality &
  Avg. \\ 
  \midrule
Tune-A-Video~\cite{tune_a_video} &
  27.12 &
  92.40 &
  20.36 &
   &
  - &
  - &
  - &
  - \\
VideoCrafter~\cite{he2022lvdm} &
  25.55 &
  88.51 &
  19.17 &
  &
  0.375 &
  0.298 &
  0.317 &
  0.330 
   \\
Text2Video-Zero~\cite{khachatryan2023text2videozero} &
  25.88 &
  92.07 &
  19.82 &
  &
  0.448 &
  0.493 &
  0.516 &
  0.486 
   \\ 
   \midrule
PAIR &
  25.53 &
  \textbf{92.47} &
  19.79 &
   &
  0.399 &
  0.402 &
  0.387 &
  0.396 \\
RewardT2VE &
  \textbf{27.55} &
  92.17 &
  20.55 &
   &
  0.451 &
  0.446 &
  0.438 &
  0.445 \\
Noah Wukong &
  27.21 &
  91.25 &
  \textbf{20.72} &
   &
  0.538 &
  0.348 &
  0.465 &
  0.450 \\
T2I\_HERO (2nd Place) &
  25.57 &
  92.27 &
  20.22 &
   &
  0.531 &
  \textbf{0.601} &
  0.564 &
  0.565 \\
CAMP (Winner) &
  26.89 &
  89.90 &
  20.71 &
   &
  \textbf{0.689} &
  0.486 &
  \textbf{0.599} &
  \textbf{0.591} \\
  \bottomrule
\end{tabular}%
}
\end{table}

\section{Team CAMP's winning method: Two-Stage Video Editing (2SVE)}
\begin{figure*}
  \centering
  \includegraphics[width=1\linewidth]{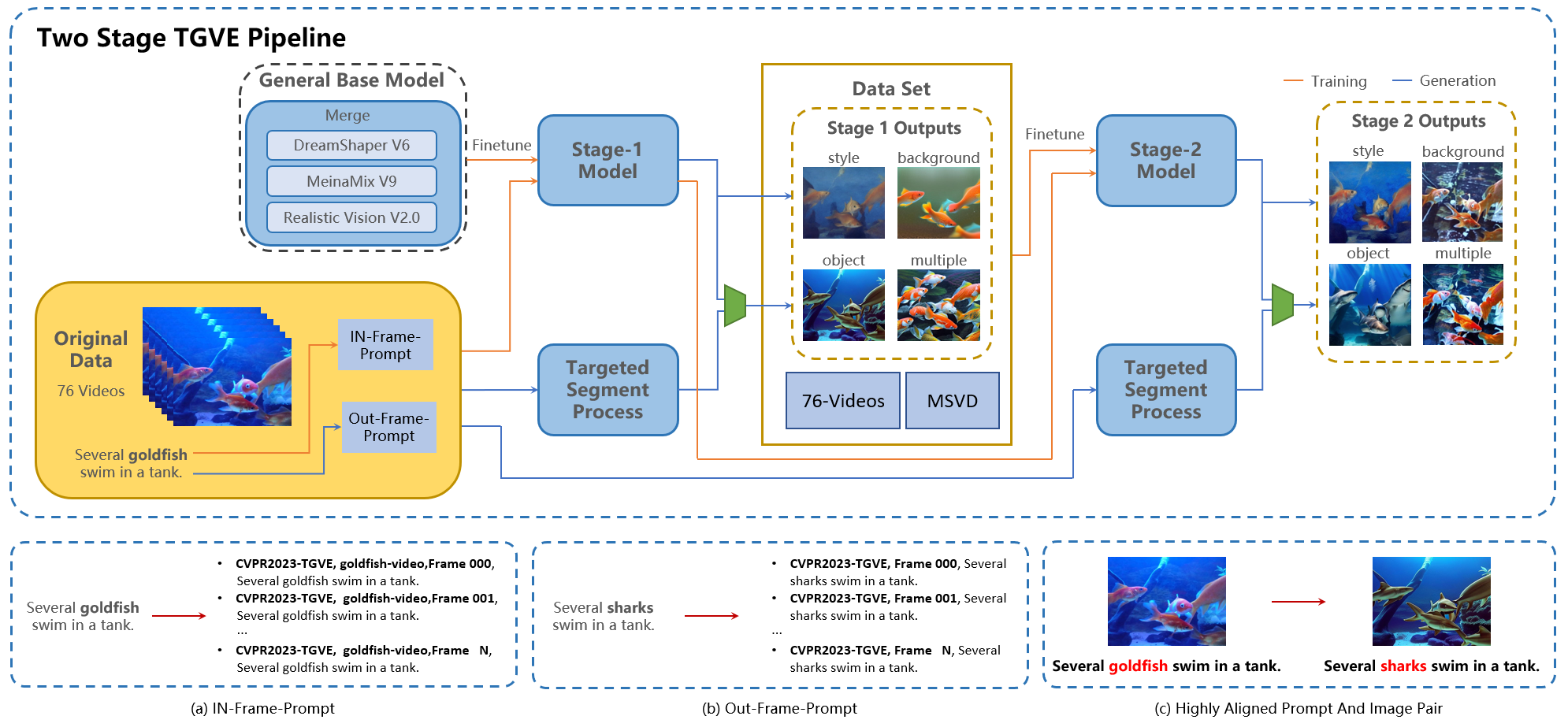}
  \caption{Pipeline of Two-Stage Video Editing (2SVE).}
   \label{fig:model}
\end{figure*}

\begin{figure}
  \centering
  \includegraphics[width=0.5\linewidth]{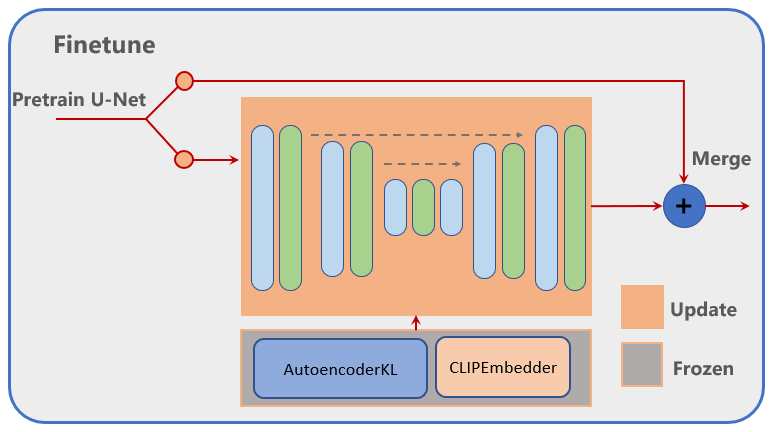}
  \caption{Overview of the finetuning stage.}
   \label{fig:finetune} 
\end{figure}

\textit{This section was written by the competition winners from Tencent Holdings Ltd and Tsinghua University. 
Figures~\ref{fig:model}--\ref{fig:multi-frame-rendering} were created by the competition winners.
}

We propose a two-stage video-to-video editing method using both text and image guidance based on diffusion models, called Two-Stage Video Editing (2SVE). The method can edit the foreground, background, and style of the input video according to the given prompt. The model structure is shown in Figure~\ref{fig:model}. In the first stage, we divide the current four tasks: style, object, background, and multiple, into two types for future processing. For the style transfer tasks without too many changes in texture or structure, according to the given prompt, we use ControlNet~\cite{controlnet} together with either LoRA~\cite{lora} or a reference-only image for the corresponding style to guide the generation (Figure~\ref{fig:style-change}). For other tasks that need to make changes to the texture or structure of the video, we finetuned the diffusion model for image generation using the video frames with their frame prompt to get our Stage 1 model. More specifically, each frame prompt contains the timestamp as well as the video name to make it unique, and the added unique identifier is called Video Prompt Anchor (VPA) (Figure~\ref{fig:model}-a, b), see Section~\ref{sec:video-prompt-anchor}. In the second stage, we further finetuned the model with both the original data and the frames generated by the Stage 1 model, hence to get our Stage 2 model. Since the output frames of Stage 1 have a highly structural alignment with the target prompts compared with the original video frames, we call these data "highly aligned prompts and images" (Figure~\ref{fig:model}-c), see Section~\ref{sec:highly-aligned-prompt-and-images} for more detail. During the generation process of the two stages, we use ControlNet~\cite{controlnet} to manipulate the frame generation. At the same time, by using both the Segment Anything Model (SAM)~\cite{segmentanything} and OpenCLIP~\cite{openclip}, combined with traditional computer vision skills including erode and dilate, we created a method called Target Segment Process (TSP) to automatically process the input frames for ControlNet (Figure~\ref{fig:targeted-segment-process}), see Section~\ref{sec:targeted-segmentation-process}. By using the Video Prompt Anchor, Multi-frame rendering\footnote{https://xanthius.itch.io/multi-frame-rendering-for-stablediffusion}, and ControlNet, we improve the coherency and consistency of video generation, see Section~\ref{sec:coherence-and-consistency}. After going through the two-stage text-guided video editing framework (2SVE), given an input video $X=\{x_0, x_1, x_2....x_N\}$, where $x_i$ represents the $i$-th input frame of the video seperating by 0.1 seconds, we finally compared the results of multiple versions from the two stages: $Y_1=\{y_{10},y_{11},y_{12}....y_{1N}\}$ and $Y_2=\{y_{20},y_{21},y_{22}....y_{2N}\}$, and select the best output as the final competition submission. Finally, by using VPA for the text control, TSP and ControlNet for the image control, multi-frame rendering for continuity control, and the highly aligned prompt and images, we ensured the alignment of the generated video and text, and preserved the coherency, consistency, and structure of the generated video.

\textbf{Finetune: } \ Many video generation models are created based on text-to-image generation models, which generate videos by adding additional time sequence modules and improving the generation capabilities along the time axis. At the same time, in order to maintain the original generation ability of the text-to-image model, fewer parameters are selected to be updated from the original model. Pix2Video~\cite{ceylan2023pix2video} uses a pre-trained structure-guided image diffusion model to perform text-guided edits on an anchor frame. Tune-A-Video~\cite{tune_a_video} leverages the pre-trained T2I diffusion models for T2V generation by updating spatiotemporal attention (ST-Attn). Video ControlNet~\cite{controlnet} does not require any training or fine-tuning of the diffusion models. Our method finetuned the entire UNet with the VAE and CLIP frozen. Finally, in order to reduce overfitting and improve the generation diversity, we merged the finetuned model with the base model with weights of 0.5 each as well. As shown in Figure~\ref{fig:finetune}, our method does not use ControlNet during the training phase. The base model for finetuning is chosen to be a weighted merge of DreamShaper\footnote{https://huggingface.co/Lykon/DreamShaper} and MeinaMix\footnote{https://civitai.com/models/7240/meinamix}, with a weight of 0.5 each.


The main advantage of our method is that the frame-by-frame video editing process can reduce the usage of computing resources and can support higher resolution and arbitrary duration video editing tasks. By adding methods such as TSP and highly aligned prompts, we provide greater control over video editing capabilities and achieve some more substantial and difficult video editing tasks. We have also tried to improve the consistency of video editing through certain methods. However, because our method is still frame-by-frame video editing, how to better improve the coherence of the generated part in the video in actual generation tasks requires further research.

\subsection{Video Prompt Anchor}
\label{sec:video-prompt-anchor}
There are already some works such as DreamBooth~\cite{dreambooth} that represent a given subject with rare token identifiers during fine-tuning a pre-trained diffusion model. Inspired by this, we added a unique identifier text based on the video prompt, namely "CVPR2023-TGVE", to represent the style and the distribution of the competition dataset, we call this the Video Prompt Anchor (VPA). By adding VPA in the generation stage of the image-based video editing, the generated results have achieved stronger stability and continuity. In the training phase, we also added video names to improve the differences in prompt text embedding between videos, but in the generation phase, we did not add this part of the information, because the video name contains object information (eg. goldfish-video), and thus will interfere with video editing results (Figure~\ref{fig:model}-a, b). By adding the Video Prompt Anchor in the generation stage, it can better generate videos that retain the original video structure including color and object texture, and reduce the randomness of single frame generation together with the help from ControlNet. The comparison of the generation effects of various methods is shown in Figure~\ref{fig:video-prompt-anchor}. The advantages of Video Prompt Anchor in maintaining the coherence and consistency of video will be introduced in Section~\ref{sec:coherence-and-consistency}. Figure~\ref{fig:continuity} for more effects of using only VPA+ControlNet to obtain video consistency and coherence at different stages with various generation methods, and Figure~\ref{fig:multi-frame-rendering} shows specifically the usage of ControlNet.

\subsection{Target Segment Process}
\label{sec:targeted-segmentation-process}
\begin{figure}
  \centering
  \includegraphics[width=1\linewidth]{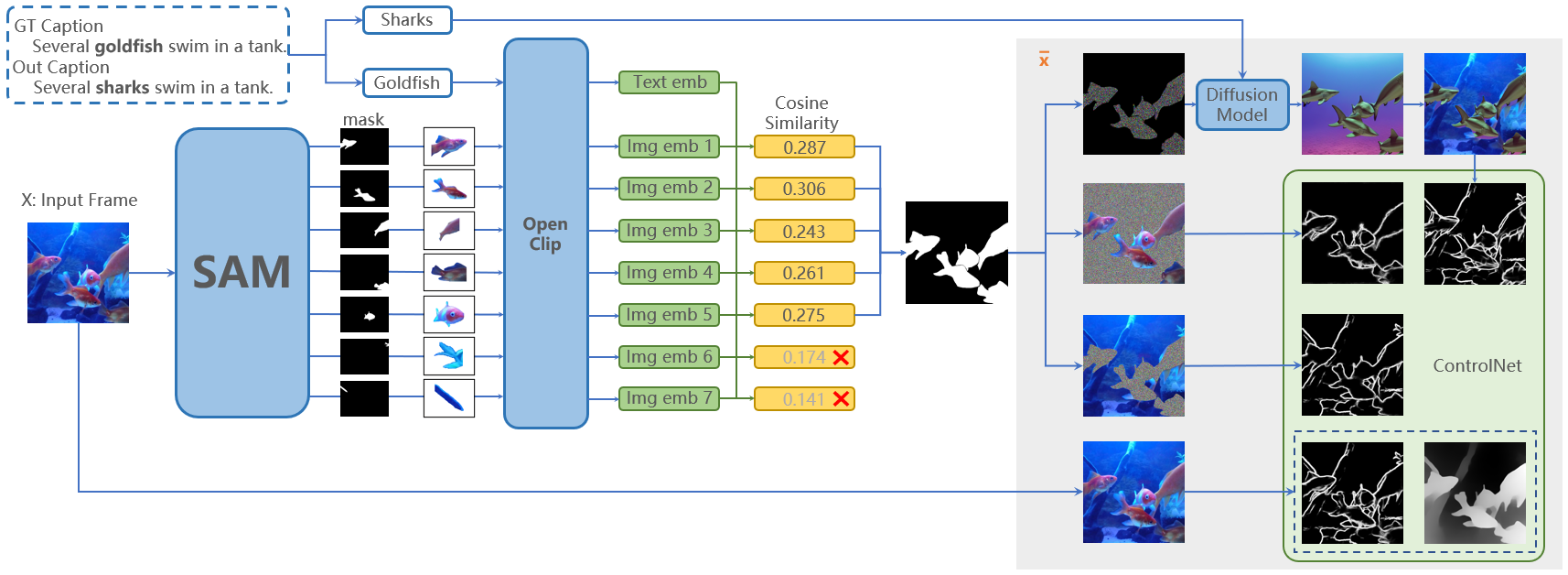}
  \caption{Illustration of Target Segment Process (TSP).}
   \label{fig:targeted-segment-process}
\end{figure}

In the video editing task, some methods usually use the input video as a reference during generation~\cite{videotovideo, fewshot}, but there are also other methods that do not rely on the input video after fully finetuned the model on the input video~\cite{tune_a_video}. Target Segment Process (TSP) is our method that comprehensively processes the input image for the ControlNet. We combine models such as Segment Anything Model, OpenCLIP, and ControlNet, together with our finetuned model, with the help of traditional image processing techniques including dilate, erode, to transfer input video frames $X=\{x_0,x_1,x_2....x_N\}$ into $\bar{X}=\{\bar{x}_0, \bar{x}_1, \bar{x}_2...\bar{x}_N\}$, as the input of the ControlNet. Thereby reducing the difficulty introduced by the large changes in texture and structure before and after the generation process. Figure~\ref{fig:targeted-segment-process} introduces the pipeline of TSP. We extract the embeddings from both the automatically extracted prompt text changes before and after editing (eg. goldfish) and the SAM segmented mask through OpenCLIP's text-encoder and image-encoder respectively, and then we calculate the similarity distance between the text and the image embedding, so as to automatically determine the target mask that needs to be processed. After getting the mask, we will try a variety of processing methods in different schemes as the input and reference for further generation, as shown in Figure~\ref{fig:targeted-segment-process-result}.

\subsection{Highly Aligned Prompts and Images}
\label{sec:highly-aligned-prompt-and-images}
\begin{figure}
  \centering
  \includegraphics[width=1\linewidth]{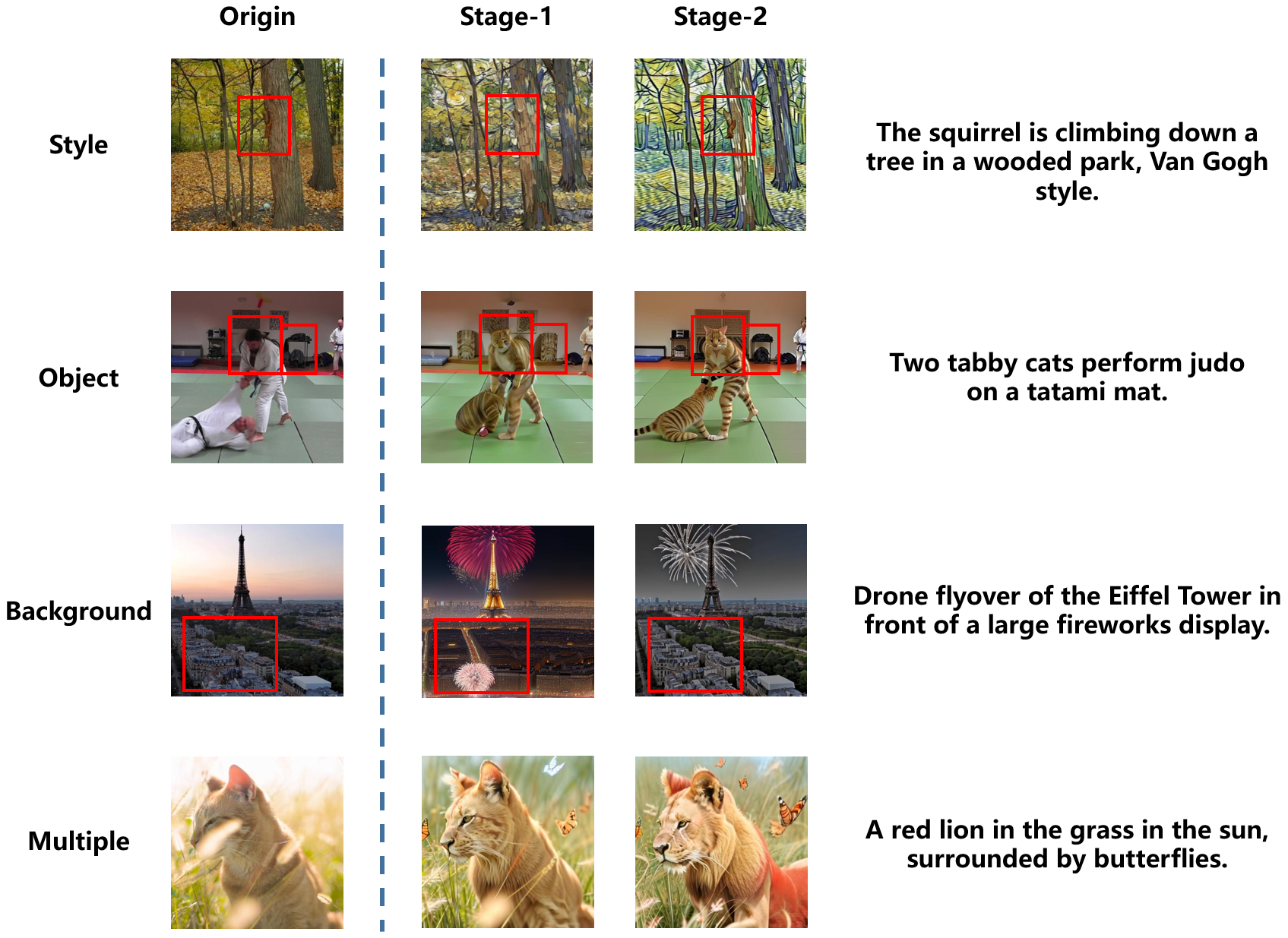}
  \caption{A comparison of the effects generated by the first stage and the second stage is shown. The second stage model has the ability to change the style under the control of text only (e.g. adding Van Gogh style). Meanwhile, the second stage can better maintain the video structure, generation quality, and the alignment of text and video. As shown in the figure, the generation of the small target of the squirrel is retained in the Style transfer task, and its consistency and coherence are well performed in consecutive frames, see also in Figure~\ref{fig:stage-1-2-compare}. In the Object task, the heads of tabby cats and items in the background are all better preserved. For the Background task, the urban structure under the Eiffel Tower is preserved as a whole, and the overall style is still consistent with the original video. At the same time, the generation of fireworks and the integration of the overall scene are more natural. Editing ability from cat to red lion has been improved in multiple changes as well.}
   \label{fig:comparison}
\end{figure}

In order to improve the model's ability to edit video frames, we add the generated data from the first stage together with the original training data to fine-tuning of the second stage model. Figure~\ref{fig:model}-c shows the original data of the goldfish video and the output of Stage 1. We think that this part of the data is highly aligned in terms of prompt and frame from the original videos, compared to other datasets (MSVD~\cite{chen2011collecting}) introduced additionally. At the same time, in order to prevent the newly added generated data from affecting the original training data, we filtered these generated data and only added the frame ID to the prompt for training, and did not use the Video Prompt Anchor as usual. Figure~\ref{fig:stage-1-output} shows some examples. Figure~\ref{fig:comparison} shows the comparison of the generation effect between the first stage and the second stage.

\subsection{The Preservation for Video Consistency}
\label{sec:coherence-and-consistency}
Many existing video processing methods align the time sequence in multiple frames, or they add the time sequence-based attention structure into the image-based diffusion model to obtain sequential information~\cite{videotovideo, align, makeavideo}. In Section~\ref{sec:video-prompt-anchor}, we introduced the original image generation effect of Video Prompt Anchor (VPA). Using VPA in combination with ControlNet, the model can effectively restore the original image, and thus can ensure the coherence and consistency of the original image. Based on this idea, our basic scheme adopts the method of first freezing our generated random seed, combining with using the output frame prompt (Figure~\ref{fig:model}-b) and putting the original video frame into the pre-trained ControlNet to guide the generation of the entire video. Because the structures of the input and output frame prompts are similar, the styles, textures, and hues between the generated frames based on VPA+ControlNet can ensure a high degree of coherence and consistency, as shown in Figure~\ref{fig:continuity}. In order to further improve stability and consistency, we propose Multi-frame Video rendering for StableDiffusion (namely MVSD for short). This method takes the generation result of the first frame and the previous frame of the video together with the original image of the current frame as the input of stable diffusion impainting, which refers to the characteristics of other positions on the canvas, namely the generated first and previous frames, to assimilate the style to the current frame. Based on this idea, we added the operation of the target segment process, and processed the inpainting mask more finely, which not only refers to the style of the surrounding frames, but also reduces the interference from the adjacent reference frames, as shown in Figure~\ref{fig:continuity}. We show the difference between VPA-MVSD and VPA-MVSD-TSP in the end-to-end system in Figure~\ref{fig:multi-frame-rendering}.

\section{Conclusions}
Text-guided video editing (TGVE) is a rapidly-improving field of Generative AI research.
However, there has been no standard benchmark for TGVE, and many TGVE models are closed-source, so it is difficult to determine which TGVE methods are the best.
To address this, we introduced the TGVE 2023 dataset, and we organized a TGVE competition at CVPR 2023.
Our competition received 5 submissions, and human evaluation showed that 2 of the submissions outperformed the baseline method.
The competition was won by Team CAMP, which developed a novel Two-Stage Video Editing (2SVE) pipeline that incorporates ControlNet, Segment Anything Model (SAM), OpenCLIP, and the MSVD dataset.
In human evalations, Team CAMP's 2SVE method outperformed the baseline 59.1\% of the time.


\printbibliography

\clearpage

\appendix

{\Large \bf
	\begin{center}
		Supplementary Material \\
	\end{center}
}

\section{Further details on Team CAMP's 2SVE method}

\begin{figure}[H]
  \centering
  \includegraphics[width=1\linewidth]{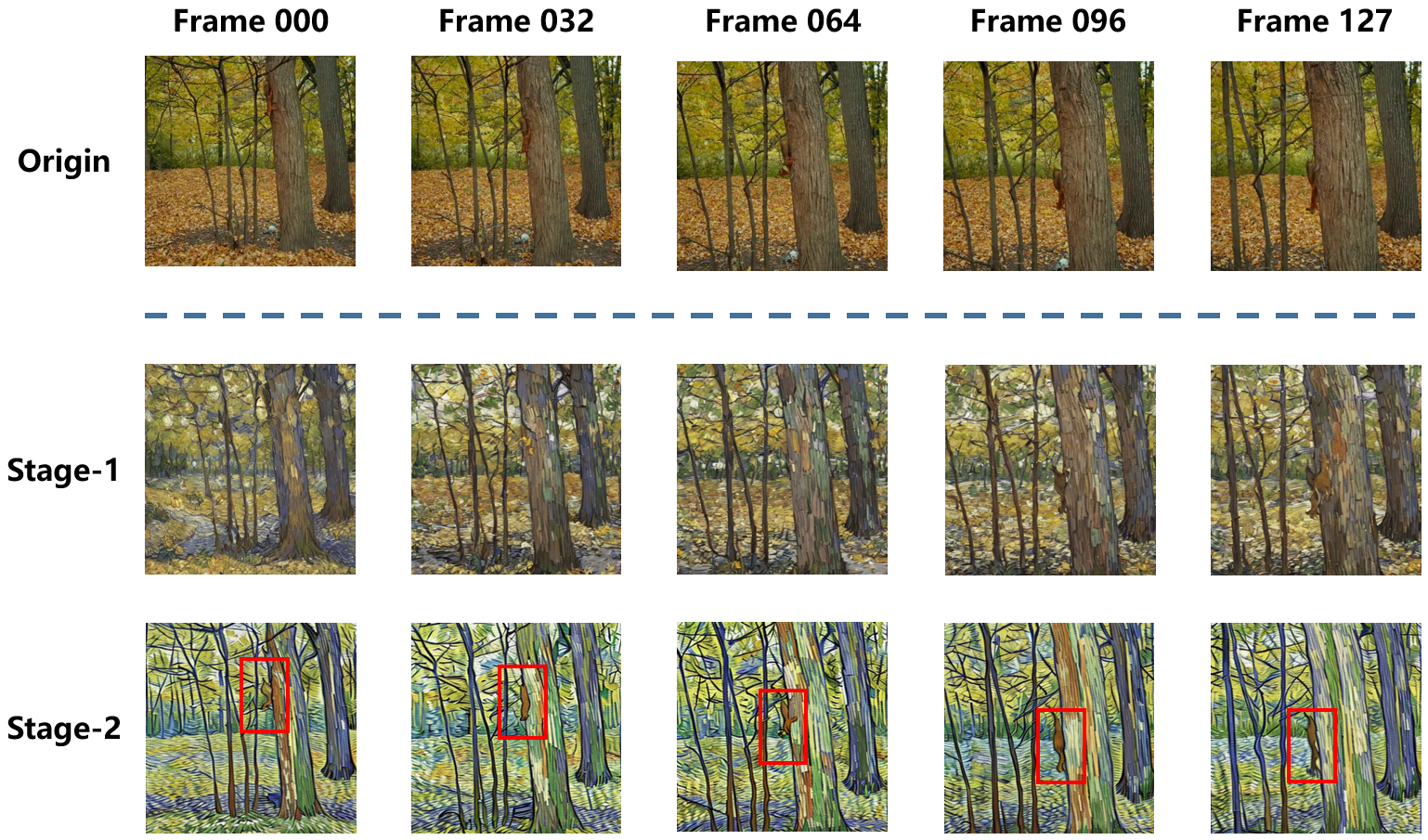}
  \caption{Taking the task of converting squirrel-climb video into Van Gogh style as an example, we compared the generation output of five frames: frames 000, 032, 064, 096, and 127 in phase 1 and phase 2 respectively. It can be seen intuitively that the results of Stage 2 have better consistency and coherence. At the same time, the main target "squirrel" can be retained in the generation result of Stage 2 as well.}
   \label{fig:stage-1-2-compare}
\end{figure}

\begin{figure}[H]
  \centering
  \includegraphics[width=1\linewidth]{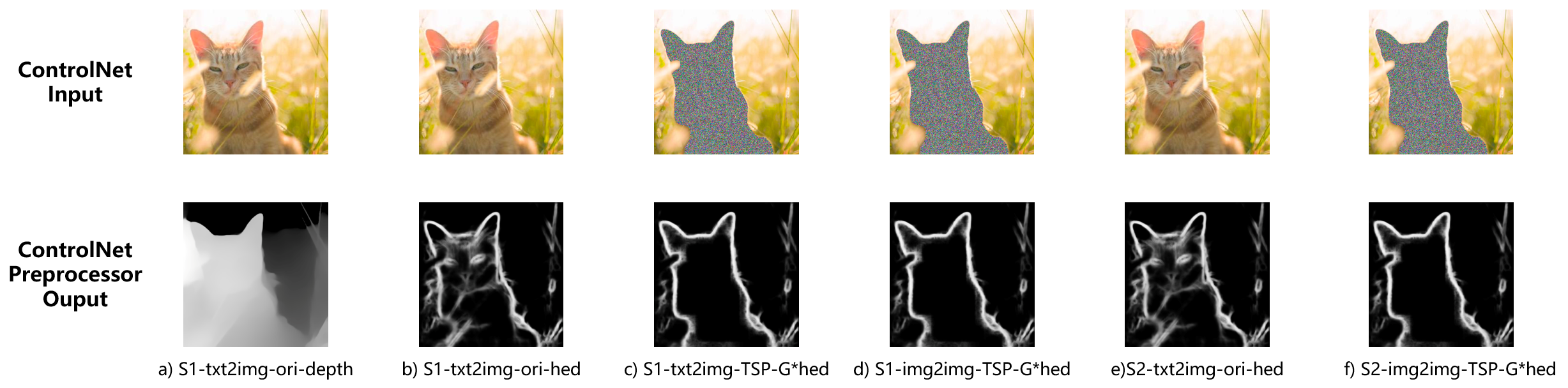}
  \caption{The various input images of ControlNet}
   \label{fig:tsp-process}
\end{figure}

\begin{figure}
  \centering
  \includegraphics[width=1\linewidth]{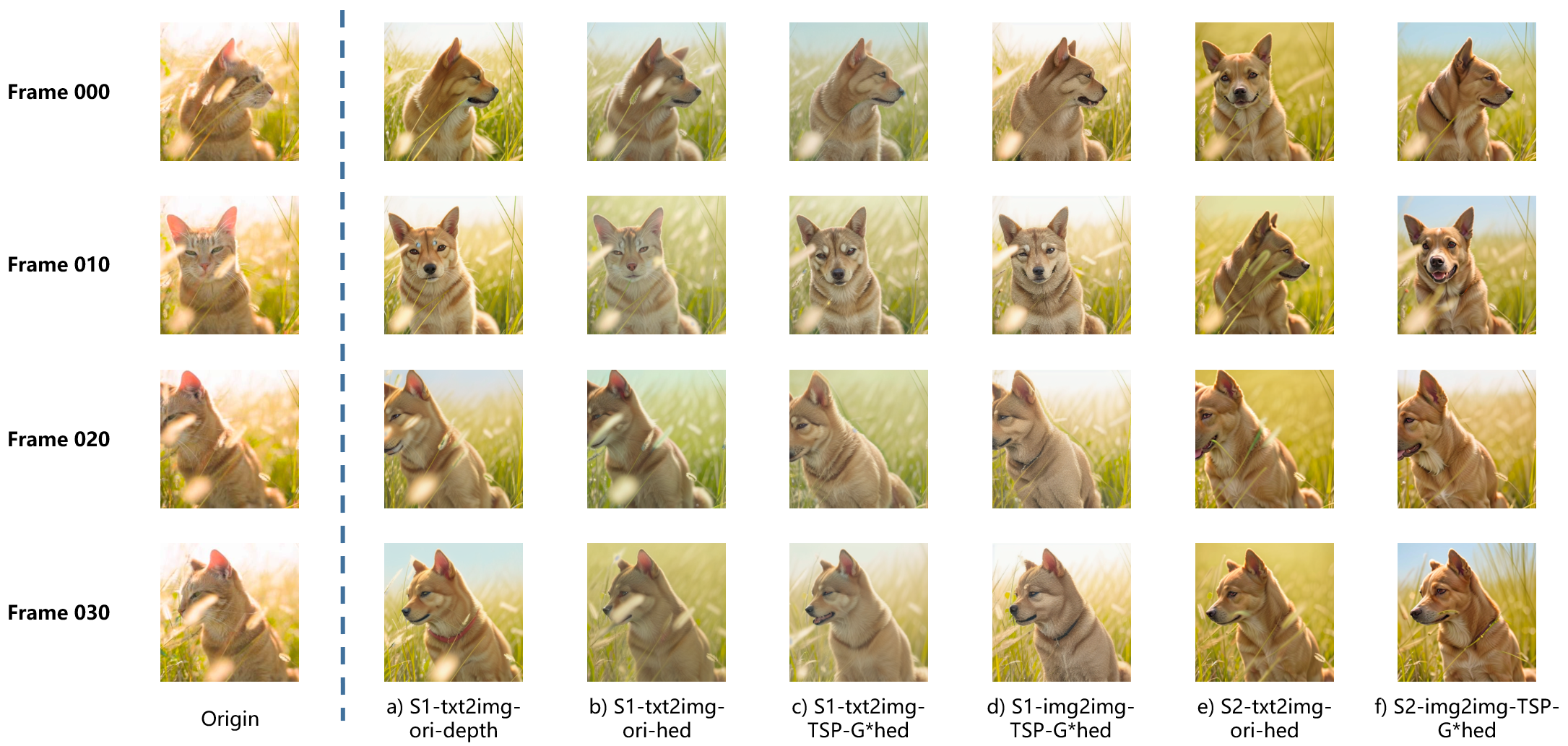}
  \caption{Under different video editing and generation methods, only VPA+ControlNet  is used to control the coherence and consistency of the video}
   \label{fig:continuity-compare}
\end{figure}

\begin{figure}
  \centering
  \includegraphics[width=1\linewidth]{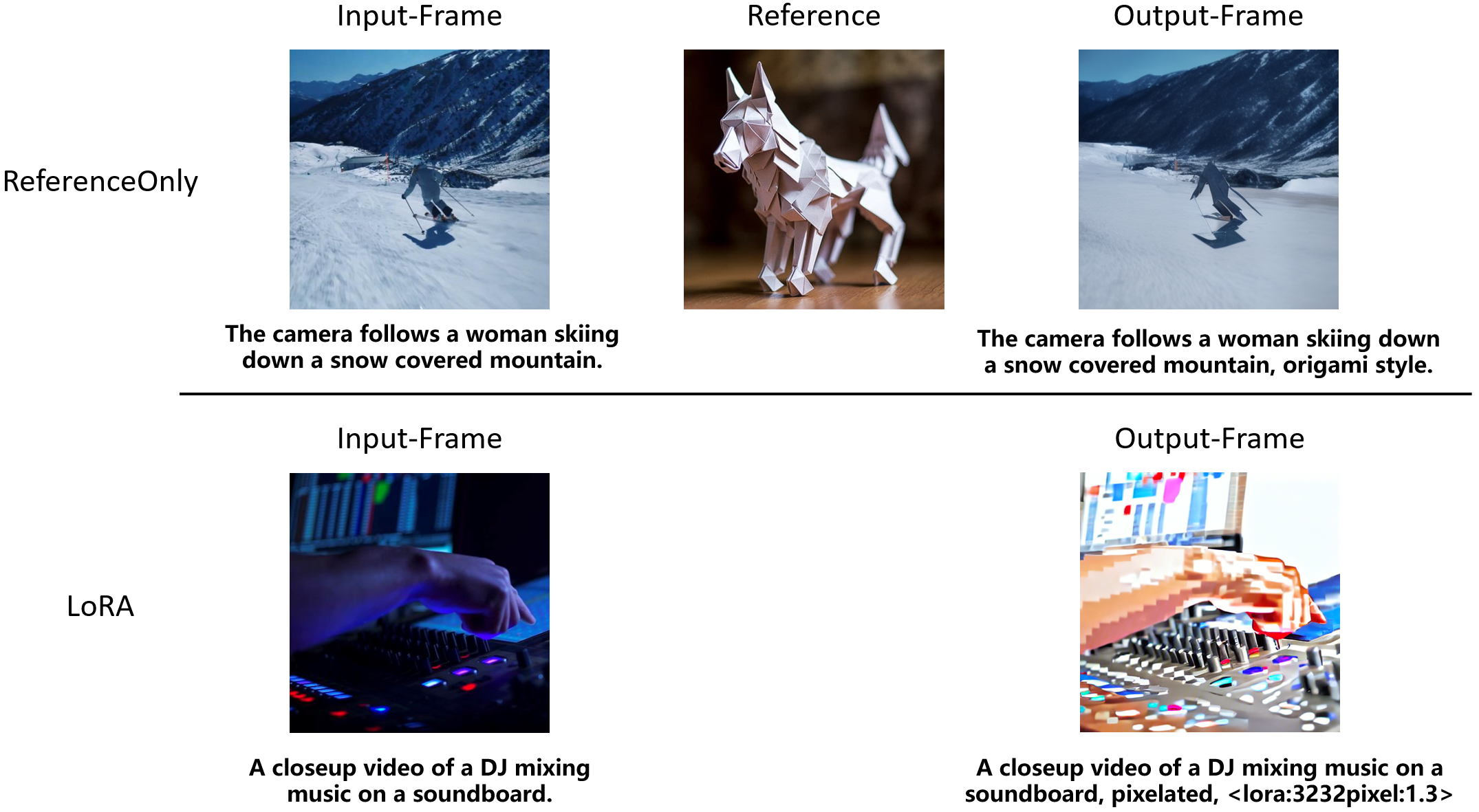}
  \caption{Comparison of the two different methods, LoRA and ReferenceOnly, that are used in the generation of video style transfer in Stage 1.}
   \label{fig:style-change}
\end{figure}

\begin{figure}
  \centering
  \includegraphics[width=1\linewidth]{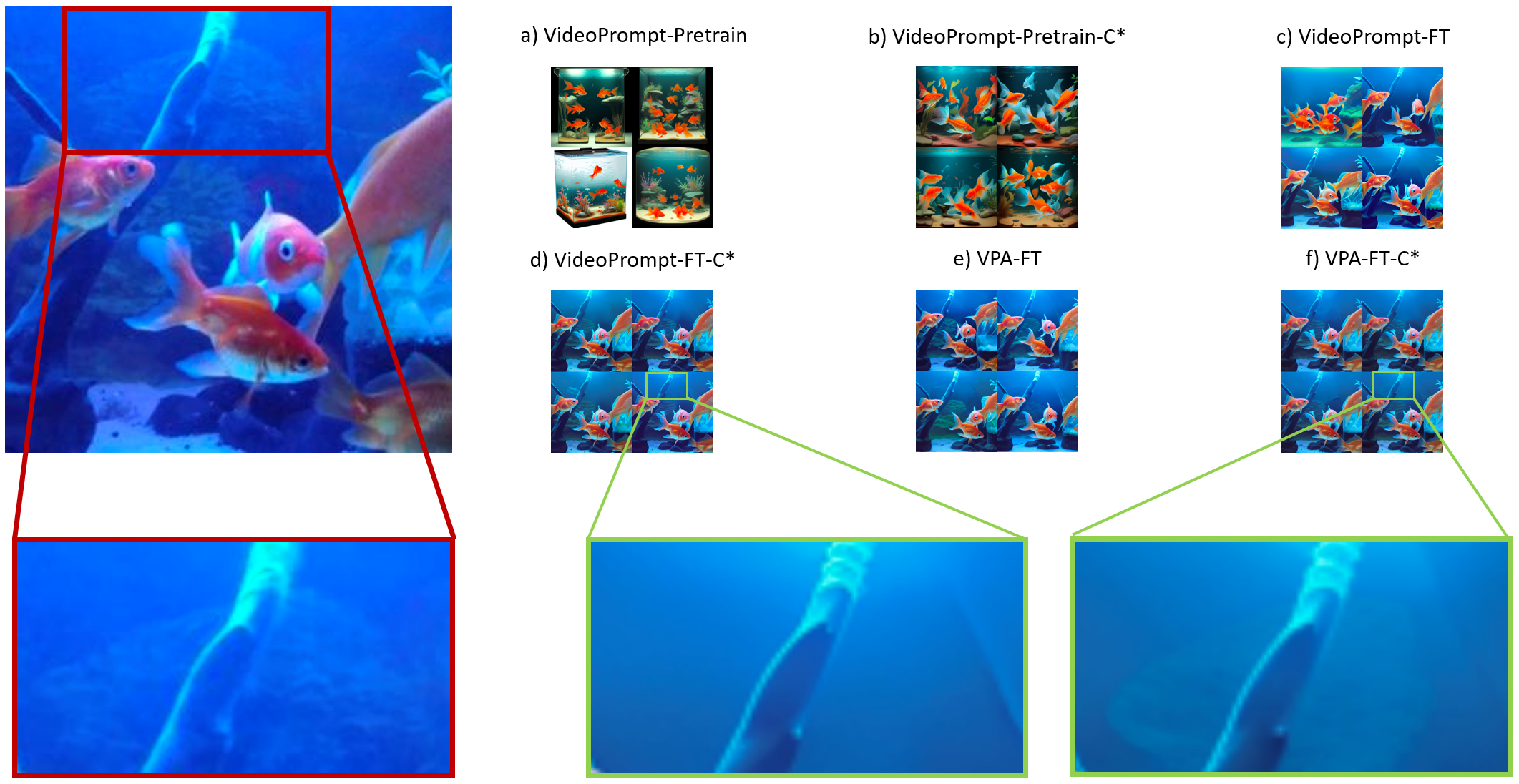}
  \caption{Comparison of the ability of VPA to reproduce and generate original video data. a) Video Prompt+Pretrain; b) VideoPrompt+Pretrain+ControlNet; c) VideoPrompt+Finetune; d) VideoPrompt+Finetune+ControlNet; e) VPA+Finetune; f) VPA+Finetune+ControlNet. The level of detail can be better expressed, as the content behind the branches in f) is more detailed than that in d).}
   \label{fig:video-prompt-anchor}
\end{figure}

\begin{figure}
  \centering
  \includegraphics[width=\linewidth]{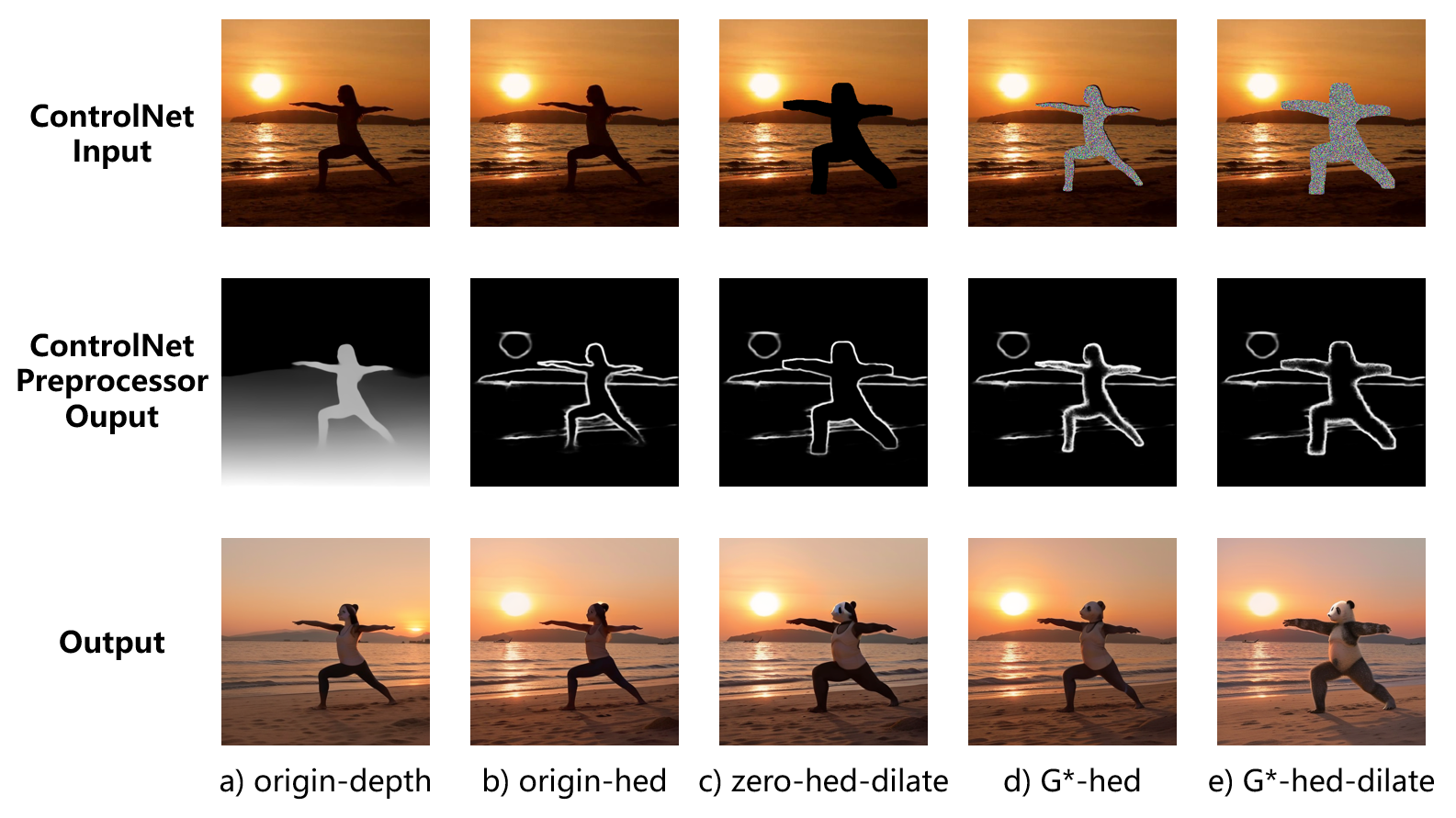}
  \caption{a) Use the depth information of the original video frame as control. b) Use the Hed softedge information of the original video frame as control. c) The result after TSP processing: the expansion operation is used after the target is segmented, and the zero value is filled. d) The result after TSP processing: the target is segmented and filled with Gaussian white noise. e) The result after TSP processing: the expansion operation is used after the target is segmented, and Gaussian white noise is used to fill it. The name of the experimental result table, the control chart filled with Gaussian white noise, in this competition, the generated details are more abundant than those filled with zero value.}
   \label{fig:targeted-segment-process-result}
\end{figure}

\begin{figure}
  \centering
  \includegraphics[width=\linewidth]{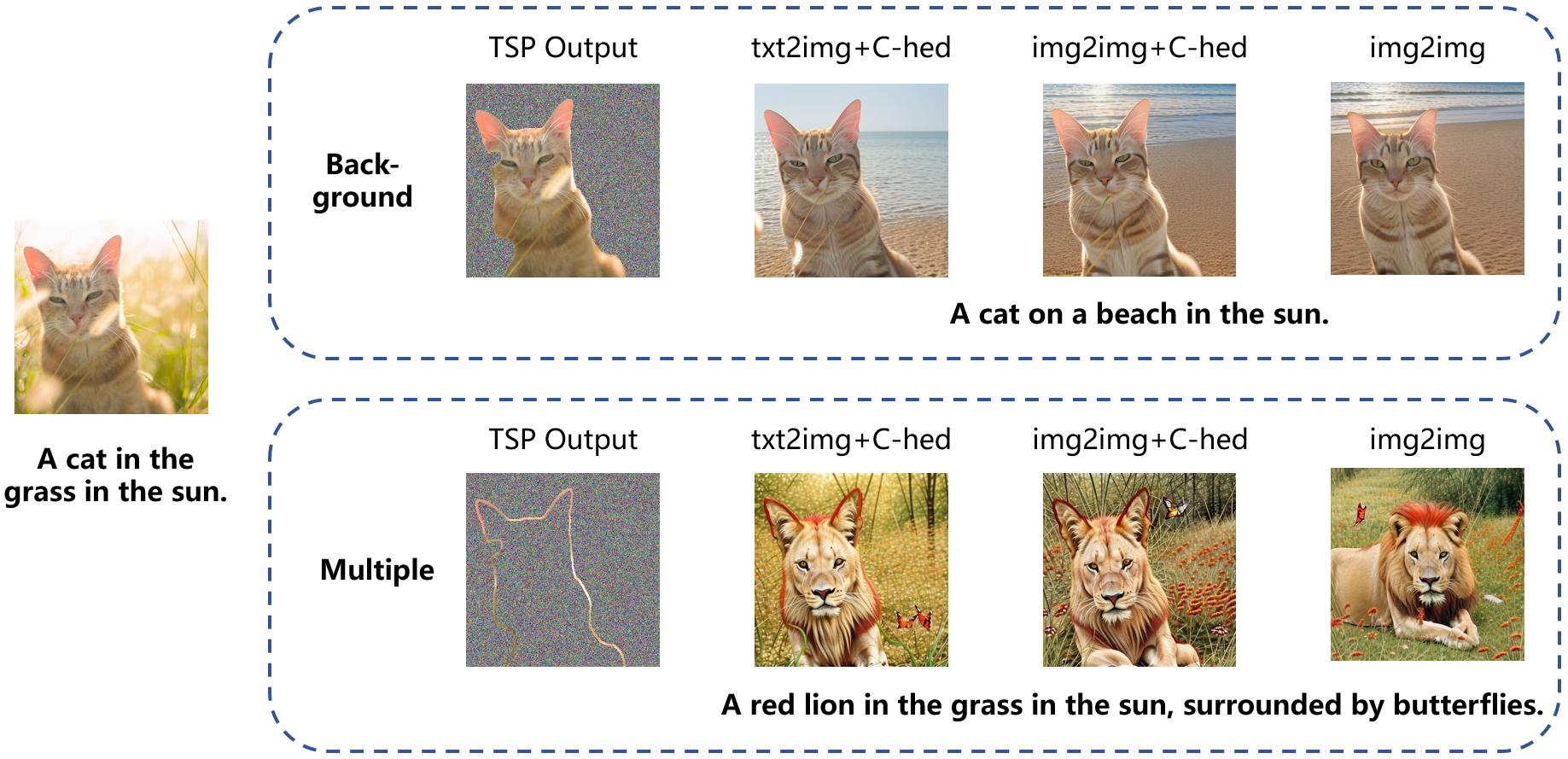}
  \caption{The generation result of Target Segment Process (TSP) under background changes and various changes. C-hed stands for ControlNet-hed control.}
   \label{fig:targeted-segment-process-result-2}
\end{figure}

\begin{figure}
  \centering
  \includegraphics[width=\linewidth]{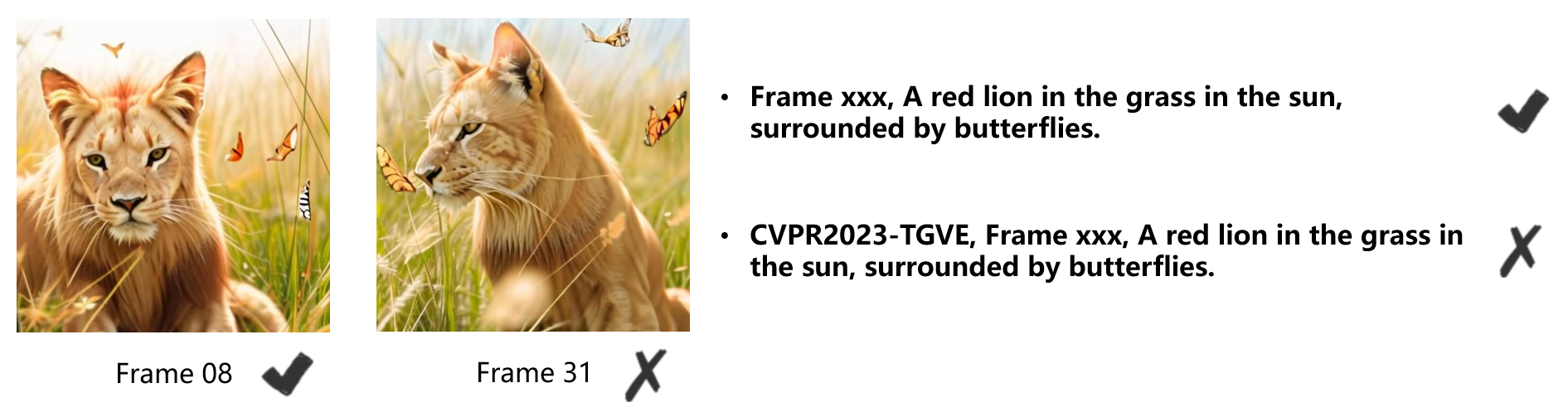}
  \caption{Stage-1-mulitple-Output of cat-in-the-sun video}
   \label{fig:stage-1-output}
\end{figure}

\begin{figure}
  \centering
  \includegraphics[width=\linewidth]{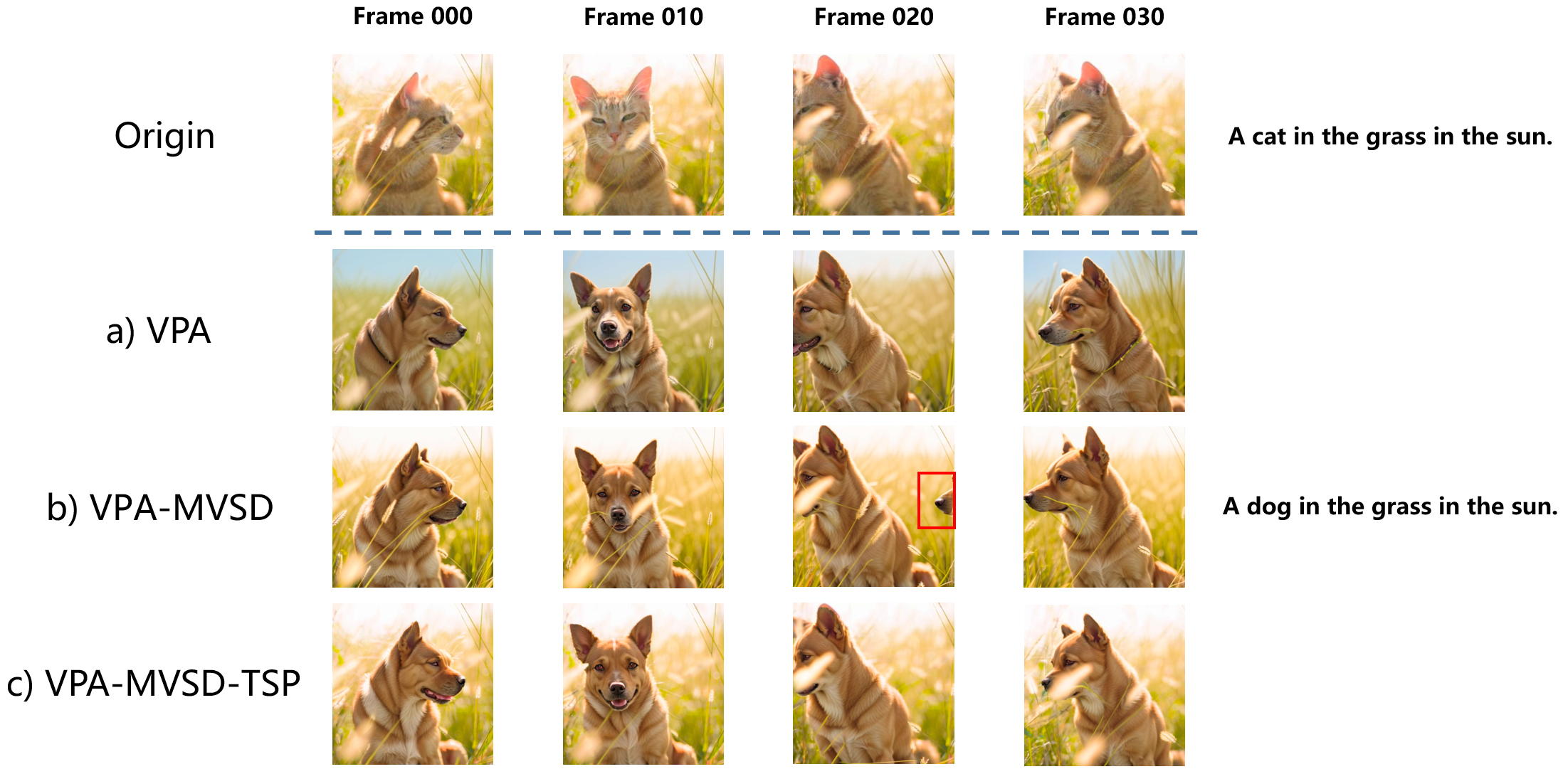}
  \caption{In this competition, we adopted three methods to ensure continuous and consistent: a) Video Prompt Anchor (VPA); b) VPA + Multi-frame-Video-rendering-for-StableDiffusion (MVSD), (see the mistake in the red box introduced by multi-frame rendering, more explained in next figure); c) VPA + MVSD + Target Segment Process (TSP).}
   \label{fig:continuity}
\end{figure}

\begin{figure}
  \centering
  \includegraphics[width=\linewidth]{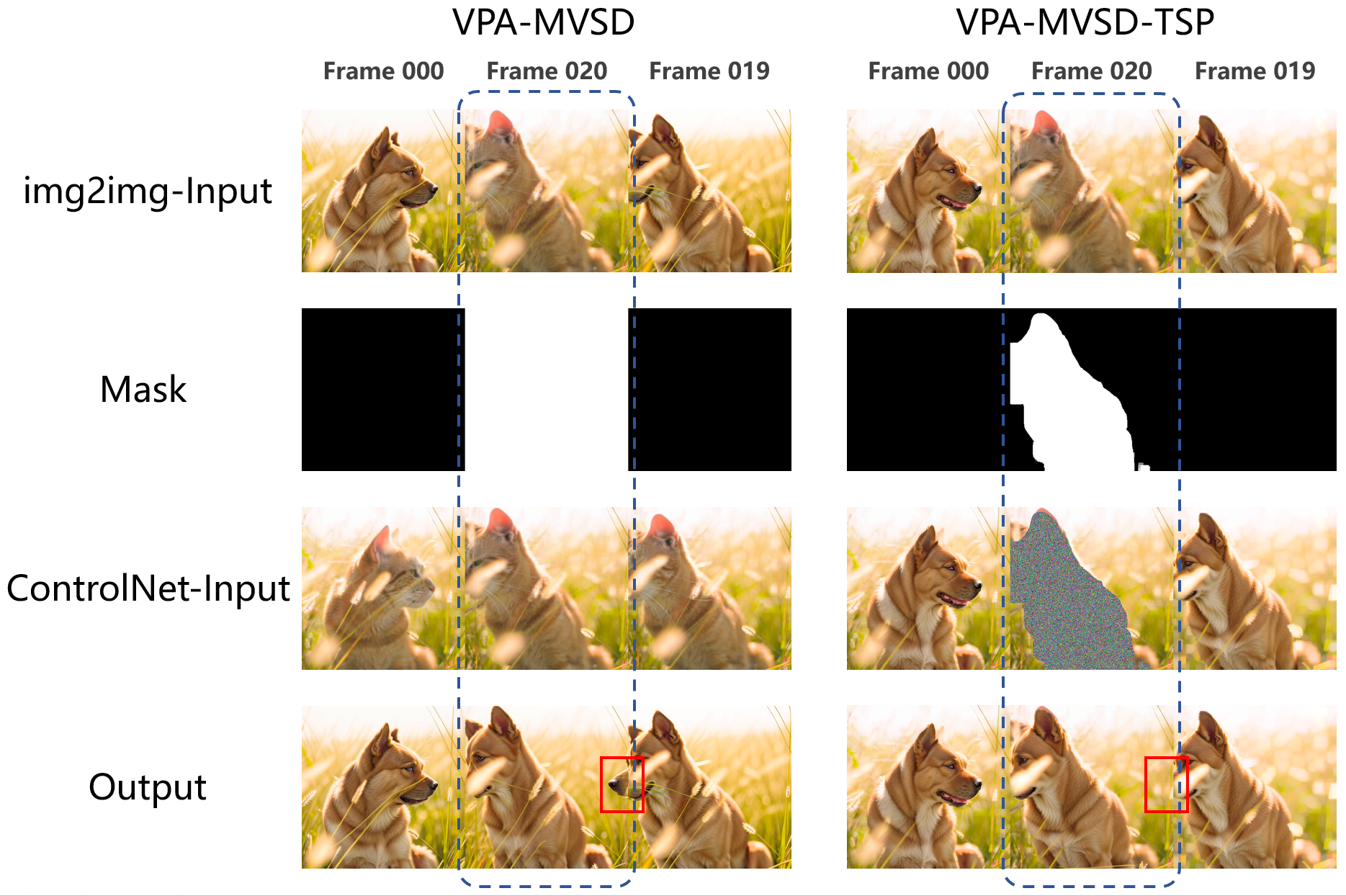}
  \caption{Comparison of the effect of standard multi-frame rendering and our MVSD+TSP. It can be seen that when generating the 20th frame of this task, by introducing the segmentation information from TSP as a more refined mask, the negative effect of the reference frame is reduced during the inpainting phase (the red box in the figure).}
   \label{fig:multi-frame-rendering}
\end{figure}

\end{document}